\documentclass[sigconf]{acmart}

\usepackage{balance}
\graphicspath{{graphics/}}
\usepackage{bm}
\usepackage{graphicx,amsmath}
\usepackage{multirow}
\usepackage{tabularx}
\usepackage{algorithm,algpseudocode}
\usepackage{algorithmicx}
\usepackage{mathrsfs}
\usepackage{color}

\newcommand{\PreserveBackslash}[1]{\let\temp=\\#1\let\\=\temp}
\newcolumntype{C}[1]{>{\PreserveBackslash\centering}p{#1}}
\newcolumntype{R}[1]{>{\PreserveBackslash\raggedleft}p{#1}}
\newcolumntype{L}[1]{>{\PreserveBackslash\raggedright}p{#1}}

\def\BibTeX{{\rm B\kern-.05em{\sc i\kern-.025em b}\kern-.08emT\kern-.1667em\lower.7ex\hbox{E}\kern-.125emX}}

\copyrightyear{2019} 
\acmYear{2019} 
\acmConference[MM '19]{Proceedings of the 27th ACM International Conference on Multimedia}{October 21--25, 2019}{Nice, France}
\acmBooktitle{Proceedings of the 27th ACM International Conference on Multimedia (MM '19), October 21--25, 2019, Nice, France}
\acmPrice{15.00}
\acmDOI{10.1145/3343031.3351024}
\acmISBN{978-1-4503-6889-6/19/10}

\begin{document}

\fancyhead{}

\title{Visual Relationship Detection with Relative Location Mining}

\author[**]{Hao Zhou}
\affiliation{%
 \institution{Shanghai Jiao Tong University}
 \city{Shanghai}
 \country{China}}
\email{zhouhao\_0039@sjtu.edu.cn}

\author{Chongyang Zhang}
\authornote{Corresonding Author. Also with MoE Key Lab of Artificial Intelligence, AI Institute, Shanghai Jiao Tong University, Shanghai, China}
\affiliation{%
 \institution{Shanghai Jiao Tong University}
 \city{Shanghai}
 \country{China}
}
\email{sunny\_zhang@sjtu.edu.cn}

\author{Chuanping Hu}
\affiliation{%
 \institution{Railway Police College}
 \city{Zhengzhou}
 \country{China}}
\email{cphu@vip.sina.com}
%
\begin{abstract}
	Visual relationship detection, as a challenging task used to find and distinguish the interactions between object pairs in one image, has received much attention recently. In this work, we propose a novel visual relationship detection framework by deeply mining and utilizing relative location of object-pair in every stage of the procedure. In both the stages, relative location information of each object-pair is abstracted and encoded as auxiliary feature to improve the distinguishing capability of object-pairs proposing and predicate recognition, respectively; Moreover, one Gated Graph Neural Network(GGNN) is introduced to mine and measure the relevance of predicates using relative location. With the location-based GGNN, those non-exclusive predicates with similar spatial position can be clustered firstly and then be smoothed with close classification scores, thus the accuracy of top $n$ recall can be increased further. Experiments on two widely used datasets VRD and VG show that, with the deeply mining and exploiting of relative location information, our proposed model significantly outperforms the current state-of-the-art.
\end{abstract}



\maketitle
\section{Introduction}
\begin{figure}[ht]
	\centering
	\includegraphics[height=8.7cm]{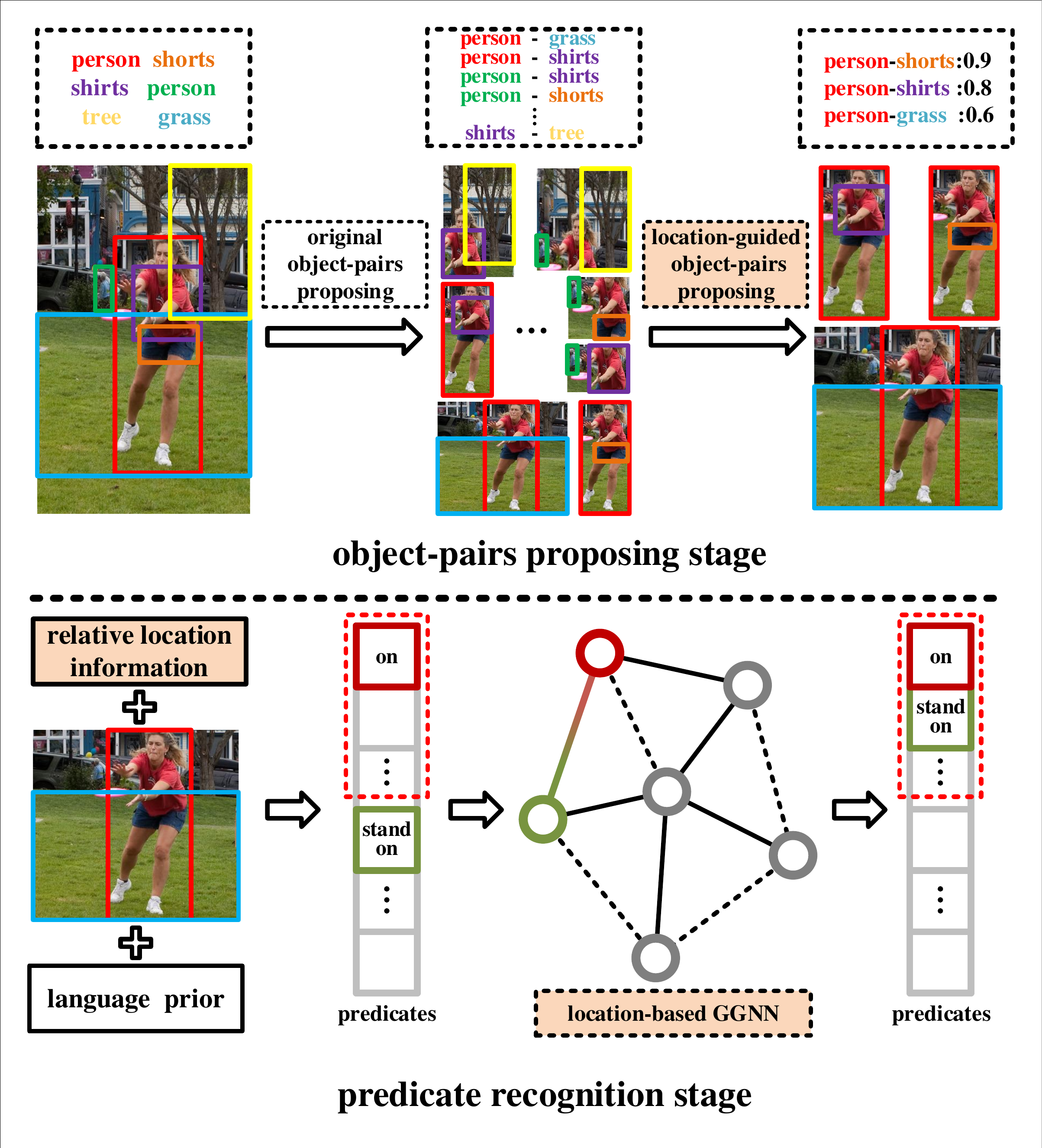}
	\caption{The illustration of our proposed framework and motivation. In the proposing stage (top row), with relative location information, some object-pairs without any spatial connection or with far relative distance, such as the pair of \textcolor[rgb]{0.22,0.70,0.31}{person} (green box)-\textcolor[rgb]{0.4,0.18,0.63}{shirts} (purple box), can be assigned with a low score and then filtered out effectively. In the predicate recognition stage (bottom row), one location-based GGNN is introduced to model relevances among predicates using relative location based similarity measuring. With this GGNN, some ambiguous and non-exclusive predicates, such as stand-on, can be smoothed and assigned proximity scores corresponding to ground-truth label, and thus the accuracy of top $n$ recall (\textcolor[rgb]{1,0,0}{red} dotted box) can be increased.}
	\label{fig:motivate}
\end{figure}
Thanks to the developing of deep learning, many deep models have achieved pretty good performances in detection\cite{ren2015faster,redmon2018yolov3} and classification\cite{wu2017adaptive,li2018shared}, even outperform human level. However, one of the further goals in computer vision is image understanding: not only recognize objects but also catch the deep semantic information in one image. Generally, visual relationships can be expressed as triplets $\left<sub-pred-ob\right>$ briefly, where $sub$, $pred$ and $ob$ mean $subject$, $predicate$ and $object$ respectively. Based on object detection, visual relationship detection attempts to distinguish the interactions $predicates$ between object-pairs. It plays an essential role in a wide range of higher level image understanding tasks, such as paragraph generation\cite{che2018paragraph}, image captioning\cite{lu2017knowing,dong2018fast}, scene graph generation\cite{xu2017scene} and visual question answering\cite{gao2018examine,wu2016ask}. Thus, visual relationship detection has received increasing attention recently.

Visual relationship detection can be divided into two stages, including object-pairs proposing stage and predicate recognition stage. Traditional methods\cite{lu2016visual,zhang2017ppr} follow the simple framework: given $N$ detection objects, $N^2$ object-pairs are proposed in object-pairs proposing stage. The main problem is that the performance of relationship models is heavily dependent on $N$. With the increasing of $N$, the combination number will grow exponentially. To avoid the problems of combinatorial explosion, one effective proposing method is to select $M$ $(M\ll N^2)$ reasonable object-pairs for the following predicate recognition. However, this kind of object-pairs proposing scheme faces one big challenge: how to select $M$ reasonable object-pairs from $N^2$ possible combinations? Some works\cite{li2017vip,yu2017visual} attempt to reserve specific object-pairs proposals based on the objectiveness scores from detection model. But there is a large gap between the higher objectiveness scores and the more meaningful object-pairs, which deteriorates performances inevitably in this proposing scheme. 

In the predicate recognition stage, one main challenge is that, many predicates are ambiguous and non-exclusive, which makes it hard to produce the most matching predictions in relationship detection task. For example, the predicates of "beside" and "near", belong to different categories but with similar meaning. Another sample is "stand on" and "on", as shown in Figure~\ref{fig:motivate}, one can say the person is standing on the grass, thus the predicate "stand on" is also an acceptable prediction compared to the ground-truth "on". Thus, those non-exclusive $predicates$ with similar spatial position should be clustered and be smoothed with close classification scores, so that several ambiguous but reasonable predictions can be produced in the inference stage. 

Considering the fact that, most visual relationships are semantic concepts defined by human beings, there are much human knowledge, or priors, hidden in them, which has not been fully exploited by existing methods. One important type of hidden knowledge is the relative location information of $\left<sub-ob\right>$ in one object-pair. In fact, many semantic relationships between object-pair are defined either according to the relative location, or closely related to the position relationship. Such as, in the triplet of "person ride bicycle" and "person wear shirt", there is one implied relative location of "on" and "overlap", respectively. That's to say, lots of semantic triplets may have a specific relative location between the pair of $\left<sub-ob\right>$: above or under, connected or without overlap, and so on. 
Thus, relative location information can offer strong inferring to find and distinguish the visual relationship of interacted object pair. 
Although there are some recent works introducing relative location information into the relation model, most of them only utilize it in one of the two stages and cannot exploit this valuable information fully.

In this work, based on above observation and motivation, we propose a novel visual relationship detection framework by deeply mining and utilizing relative location of object-pairs in both the two stages. Firstly, one location-guided Object-pairs Rating Module(ORM) is proposed and combined with object detection module to select as well as sort valid object-pairs proposals in the object-pair proposing stage; Moreover, one location-based Gated Graph Neural Network(GGNN) is introduced to mine and measure the relevance of $predicates$ using relative location. 
With the location-based GGNN, those non-exclusive $predicates$ with similar relative location are clustered in the training stage. Then, some ambiguous $predicates$ are smoothed and assigned close classification scores, so that the accuracy of top $n$ recall can be increased. 
Experiments on two widely used datasets Visual Relationship Detection(VRD)\cite{lu2016visual} and Visual Genome(VG)\cite{krishna2017visual,zhang2017visual} show that, with deeply exploiting of relative location information, our proposed framework significantly outperforms current state-of-the-art methods.

\begin{figure*}[!th]
	\centering
	\includegraphics[width=17cm]{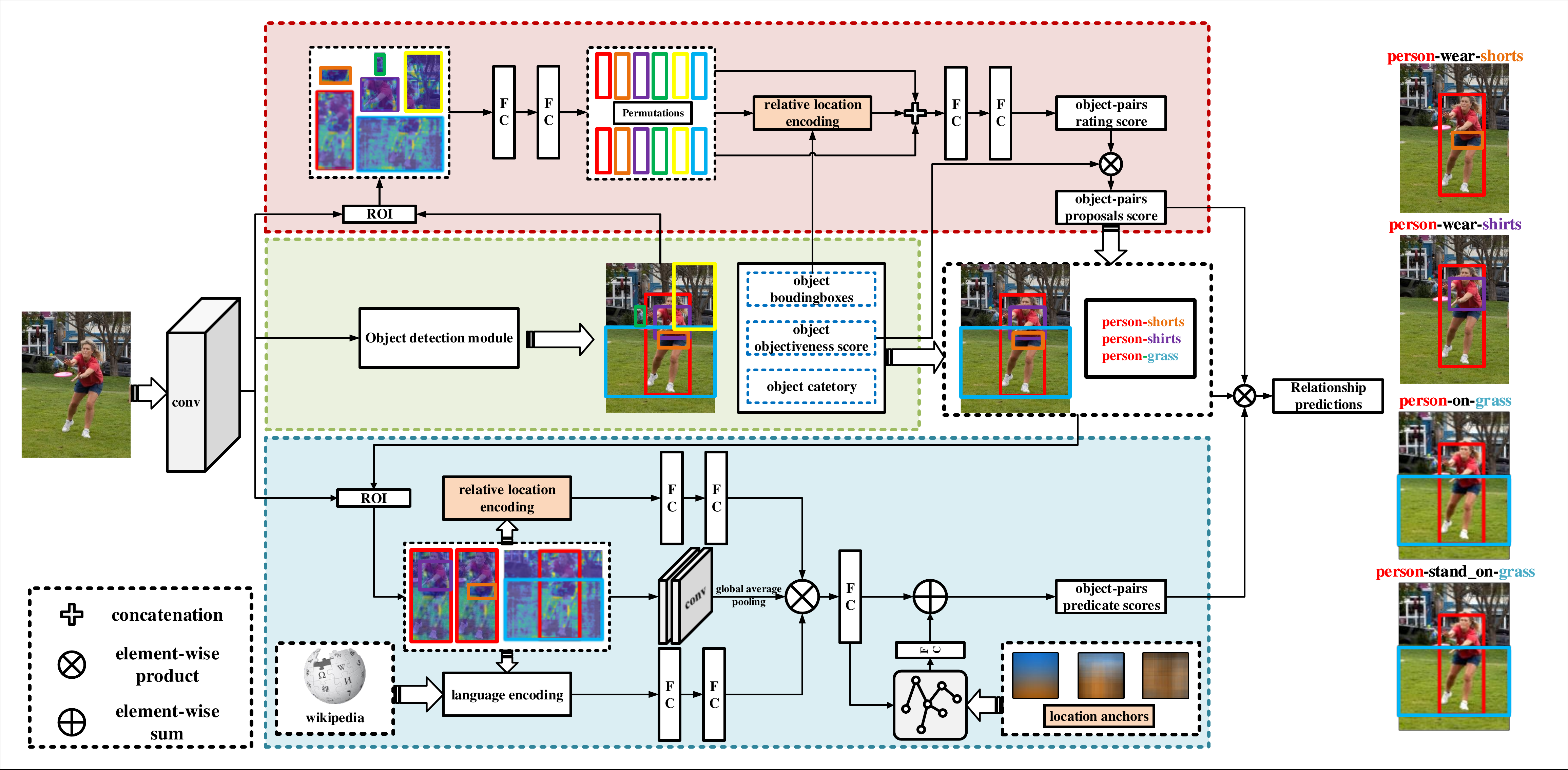}
	\caption{The detailed structure of our proposed visual relationship detection framework. From top to down, \textcolor[rgb]{1,0.5,0.5}{Red}, \textcolor[rgb]{0.56,0.85,0.30}{green}, \textcolor[rgb]{0.1,0.75,0.82}{blue} dot boxes are the location-guided object-pairs rating module, the object detection module and the predicate recognition module with location-based GGNN, respectively. }
	\label{fig:2}
\end{figure*}

\section{Related work}
$Visual\, Relationship\, Detection$. In the early years, \cite{sadeghi2011recognition,simonyan2014very} distinguish relationships as phrases, which have poor generalizations because of the large scale of relationships. Recently, most methods\cite{lu2016visual,zhang2017visual,yu2017visual,zhuang2017towards} are developed to detect relationships expressed as triplets. One detector focuses on distinguishing $predicates$, and another one focuses on distinguishing $subjects$ and $objects$. In this way, given $N$ objects and $K$ predicates, only $N+K$ classifiers are needed to detect $N^2K$ visual relationship triplets. In fact, visual relationship detection has a wide range of concepts. There are also many works focusing on a specific type of relationships. For example, \cite{galleguillos2008object} attempts to distinguish spatial relationships and \cite{rohrbach2013translating,desai2012detecting} only focus on human-object interaction. In \cite{lu2014online}, visual similarities are exploited and fused to classify objects online. In our work, we focus on more general visual relationship detection. 

In the object-pairs proposing stage, \cite{li2017vip} proposes a triplet proposal with NMS, based on the product of objectiveness scores, to remove redundant object-pairs. However, there exists a gap between higher objectiveness scores and more meaningful object-pairs obviously. Thus, we construct an Object-pairs Rating Module to model the probability of object-pairs directly.
Noteworthy, although \cite{zhang2017relationship} proposes "relationship proposal networks" based on similar motivation, there are many differences between ORM and their approach. Different from generating visual scores, spatial scores separately, and then re-scoring final proposing scores, our ORM combines visual and relative location information and produces final object-pairs rating scores directly. Moreover, their work only focuses on the object-pairs proposing task and ignores the influences of objectiveness scores in the object-pairs proposing stage. The object-pairs proposing scheme of our network takes both object detection module and object-pairs rating module into considerations and sorts the reserved object-pairs proposals, which achieves significant improvements in our experiments.

In the predicate recognition stage, the basic method \cite{lu2016visual} takes union regions of $subject$ and $object$ as inputs and adds language priors to preserve alignments with human perception. Considering the fact that visual features provide limited knowledge for distinguishing the $predicates$, many works focus on introducing different modal features into predicate recognition stage. For example, \cite{zhang2017visual,yu2017visual,zhuang2017towards,zhou2019visual} prove that language prior and location information denoting categories and location of object pairs are effective to improve the performances of visual relationship recognition. However, comparing to language prior, relative location information, as strong inferring to relationships, are not fully exploited in those works. 

$Object \, Detection$. Recently, Fast-RCNN\cite{girshick2015fast}, Faster-RCNN\cite{ren2015faster}, SSD\cite{liu2016ssd} and YOLO\cite{redmon2018yolov3} are used to improve object detection performances. Generally, the object detection module contains two components, region proposal and classification networks. To brief expression, we regard two components as a whole, called object detection module in our paper. In addition, we follow Faster-RCNN and not change the structure of object detection module, thus this module can be exchanged easily.  

$Graph \, Neural \, Network$. To model the interaction among different $predicates$, our $predicate$ recognition module is based on the Gated Graph Neural Network\cite{li2015gated}. Comparing to traditional gated recurrent models, Graph Neural Network needs a certain structure predefined through external knowledge as a prior. There are some works\cite{woo2018linknet,yang2018graph,cui2018context,lu2014online} constructing graph networks based on objects. From different motivation, we propose a novel predicate-specific graph network in the visual relationship detection task.  

\section{our approach}
\subsection{Overall}
The overall framework of our proposed relationship detection network is illustrated in Figure~\ref{fig:2}, which consists of three interconnected modules, including object detection module (green box), location-guided object-pairs rating module (red box) and predicate recognition module with location-based GGNN (blue box). Firstly, given one image, object bounding-boxes, objectiveness scores and object categories are produced from the object detection module. Then in location-guided object-pairs rating module, visual representations and encoded relative location information are used to output the rating scores that denote the probabilities of object-pairs interconnected. With the object-pairs proposals scores that combine rating scores and objectiveness scores, specific object-pairs proposals are reserved and sorted. Finally, the union visual feature maps, language encoding and relative location encoding are fed into our proposed predicate recognition module with location-based GGNN to distinguish $predicates$.

\subsection{Location Guide Object-pairs Rating}
Location-guided object-pairs rating module shares the same backbone with object detection module. Following previous mentioned motivation, the probabilities of object-pairs interconnected are measured by visual features and relative location information fusion. Visual features are extracted through ROI-Align\cite{he2017mask} pooling operation on the feature maps. To maintain a consistent dimension and release the computation resource, we set two FC-layers after ROI-Align pooling to get the representation of visual features. Finally, the visual representation vector denotes as $\mathcal{R}_{vis}\in\mathbb{R}^{n}$, where $n$ is the visual representation dimension as a hyperparameter in our experiment. 

Same with visual representations, relative location information is encoded with generated bounding-boxes from object detection module. Let $[x_{sub},y_{sub},w_{sub},h_{sub}]$ and $[x_{ob},y_{ob},w_{ob},h_{ob}]$ denote the coordinates of two bounding boxes respectively. $sub$ ($ob$) means $subject$ ($object$), $(x,y)$ denotes the upper left corner, and $w,h$ are the width and height. Considering that most relationships exist close relevance with both respective position and mutual position of object-pairs, relative location representation is encoded with both respective position and their mutual position in our work. Given a single bounding box, let $W_u, H_u$ and $S_u$ denote the width, height and area of object-pairs union region. Therefore, the respective position is represented as $[\frac{x}{W_{u}},\frac{y}{H_u},\frac{x+w}{W_u},\frac{y+h}{H_u},\frac{S}{S_u}]$, and mutual position is represented as $[\frac{x_{sub}-x_{ob}}{w_{ob}},\frac{y_{sub}-y_{ob}}{h_{ob}},\log\frac{w_{sub}}{w_{ob}},\log\frac{h_{sub}}{h_{ob}}]$. Then, L2-normalization is applied to get the final relative location representation encoding $\mathcal{R}_{loc}(sub,ob)\in\mathbb{R}^{14}$. 

In order to introduce location information into our object-pairs rating module, we concatenate $\mathcal{R}_{vis}$ and $\mathcal{R}_{loc}$ and get the representation of object-pairs as input for object-pairs rating module

\begin{equation}
\mathcal{R}_{orm}(sub,ob)= [\mathcal{R}_{vis}(sub),\mathcal{R}_{vis}(ob),\mathcal{R}_{loc}(sub,ob)].
\label{equ:1}
\end{equation}

Let $\bm s_{orm}$ be the object-pair rating score from our object-pairs rating module, which denotes the probability of the object-pair interconnected. Therefore, $\bm s_{orm}$ is defined as

\begin{equation}
\begin{aligned}
&\bm h_{orm}= f(\mathcal{R}_{orm},\bm \varTheta),\\
&\bm s_{orm}= \frac{1}{1+e^{-\bm h_{orm}}},
\end{aligned}
\label{equ:2}
\end{equation}
where $f(\cdot)$ is an output network implemented by two FC-layers, and $\bm \varTheta$ are the parameters.

Let $\mathcal{B^*} = \{\langle b^{*,1}_{sub},b^{*,1}_{ob}\rangle,\langle b^{*,2}_{sub},b^{*,2}_{ob}\rangle,\cdots \}$ denote a set of object-pairs' bounding-boxes in the training data. Due to the missing annotations for ORM, the ground-truth labels for object-pairs in $\mathcal{B^*}$ are assigned as 1, otherwise, as 0. $b_i\cap b_j$, $b_i\cup b_j$ denotes the intersection area, union area between bounding-boxes $b_i$ and $b_j$, respectively. In the training stage, the triplet IoU score computed for one object-pairs proposal $\left<b_{sub},b_{ob}\right>$ is defined as
\begin{equation}
Tri\_IoU_{\left<b_{sub},b_{ob}\right>}= \max \{\frac{b_{sub}\cap b^{*,1}_{sub}}{b_{sub}\cup b^{*,1}_{sub}} \cdot \frac{b_{ob}\cap b^{*,1}_{ob}}{b_{ob}\cup b^{*,1}_{ob}},\cdots \}.
\label{equ:3}
\end{equation}

Based on the triplet IoU scores, the labels (denoted as $y_{orm}$) of object-pairs proposals are assigned as 1 (up to $thresh\_high$) or 0 (below to $thresh\_low$). Rest proposals that mean existing ambiguity are ignored in the training stage. The loss for object-pairs rating module is

\begin{equation}
\mathcal{L}_{orm}= \frac{1}{N}\sum_{n=1}^N[y_{orm}^n\cdot \log{\bm s_{orm}^n}+(1-y_{orm}^n)\cdot \log{(1-\bm s_{orm}^n)}],
\label{equ:4}
\end{equation}
where $N$ is the batch size. 

\subsection{Rating Scores and i-NMS based Object-pair Proposing}
The scheme of our object-pairs proposing takes both object detection module and our proposed location-guided object-pairs rating module into consideration. Given each object-pair $\langle b_i,b_j\rangle$ for one image, the final object-pairs proposal score is defined as
\begin{equation}
\tilde{\bm s}(sub,ob|\langle b_i,b_j\rangle)= \bm s_{orm}(sub,ob)\cdot \hat{P}(sub|b_i)\cdot \hat{P}(ob|b_j),
\label{equ:5}
\end{equation}
where $\hat{P}$ is the objectiveness score from object detection module. Inspired by greedy NMS\cite{girshick2014rich} and triplet NMS\cite{li2017vip}, shown in Algorithm~\ref*{alg:1}, object-pairs proposing scheme is based on rating scores and improved NMS(i-NMS). 

\begin{algorithm}[!htp]
	\caption{Object-pairs Proposing Scheme based on Rating Scores and i-NMS}
	\label{alg:1}
	\begin{algorithmic}[0]
		\Require
		$\mathcal B$: object detection boxes;
		$\mathcal C$: corresponding object categories;
		$N_o$: object-pairs numbers;
		$N_t$: threshold for NMS;
		\Ensure
		$\mathcal D$: object-pairs proposals;
		$\mathcal S$: object-pairs proposals scores;
		\State initial $\mathcal D \leftarrow$ \{ \}, $\mathcal S \leftarrow$ \{ \}, $ \widetilde{\mathcal B} \leftarrow$ \{ \} and $ \widetilde{\mathcal S} \leftarrow$ \{ \};
		\For {$b_i,b_j $ in $\, \mathcal B$ and $i\neq j$}
		\State object-pairs proposal score $\widetilde{\mathcal S} \leftarrow \widetilde{\mathcal S} \cup \tilde{\bm s}(sub,ob|\langle b_i,b_j\rangle)$;
		\State corresponding object-pairs boxes  $\widetilde{\mathcal B} \leftarrow \widetilde{\mathcal B} \cup \langle b_i,b_j\rangle$;
		\EndFor
		\State $\backslash\backslash$i-NMS
		\While {$\widetilde{\mathcal B}\neq empty$ and $|\mathcal D|\leqslant N_o$}
		\State $\tilde{\bm s}(sub,ob|\langle b_m,b_n\rangle) \leftarrow \rm max \, \widetilde{\mathcal S}$;
		\State $\mathcal D \leftarrow \mathcal D  \cup \langle b_m,b_n\rangle$, $\mathcal S \leftarrow \mathcal S \cup \tilde{\bm s}(sub,ob|\langle b_i,b_j\rangle)$; 
		\State $\widetilde{\mathcal B} \leftarrow \widetilde{\mathcal B} - \langle b_m,b_n\rangle$;
		\For {$\langle b_i,b_j\rangle $ in $\, \widetilde{\mathcal B} $}
		\If { $IoU(b_m,b_i)\ast IoU(b_n,b_j) \geqslant N_t $ and \\ \qquad \qquad \;$c_m = c_i \& c_n = c_j$}
		\State  $\widetilde{\mathcal B} \leftarrow \widetilde{\mathcal B} - \langle b_i,b_j\rangle$, $\widetilde{\mathcal S} \leftarrow \widetilde{\mathcal S} - \tilde{\bm s}(sub,ob|\langle b_i,b_j\rangle)$;
		\EndIf
		
		\EndFor
		
		\EndWhile
	\end{algorithmic}
\end{algorithm}

\subsection{Predicate Recognition using GGNN}
In visual relationship detection, predicate recognition module is the key component, which attempts to recognize the interactions given the union regions of object-pairs. In order to enhance the distinguish capability, multiple modal feature fusion and location-based GGNN is developed in this work, shown the blue box in Figure~\ref{fig:2}.
\subsubsection{Multiple Modal Features Fusion}

Inspired by previous works using multiple features or priors fusion, one enhanced multiple modal features fusion, by which not only visual features, language prior, but also relative location information, are integrated together to achieve more powerful representation ability.  

In the predicate recognition stage, union visual features are extracted as visual relationship features. Two convolutional layers are employed after ROI-Align pooling to encode the visual interactions of object-pairs. To minimize the numbers of parameters as well as retain the ability of networks, global average pooling\cite{zhou2016learning} is implemented after the convolution layers. Finally, the visual features are represented as $\mathcal{R}_{vis}(sub,ob)\in\mathbb{R}^{m}$.

Language prior provides auxiliary information to infer relationships. In our work, we encode language prior through concatenating corresponding word2vec \cite{mikolov2013distributed} embedding of $subject$ and $object$. The word2vec is trained with the whole Wikipedia and mapped into 300 dim each word. Then, L2-normalization is applied on each word2vec embedding. The language representations $\mathcal{R}_l(sub,ob)\in\mathbb{R}^{600}$ are encoded as

\begin{equation}
\mathcal{R}_l(sub,ob) =[word2vec(sub),word2vec(ob)] .
\label{equ:6}
\end{equation}

To simplify the network, location information adopts the same relative location encoding defined in previous section $\mathcal{R}_{loc}(sub,ob)$. To fuse the three modal features, two separate FC-layers are implemented to map language representation and relative location representation into the same space with visual representation. The final relationship representation $\mathcal{R}_{prm}\in\mathbb{R}^{m}$ for predicate recognition module is defined as
\begin{equation}
\mathcal{R}_{prm} =\mathcal{R}_{vis}\odot f(\mathcal{R}_l, \varTheta_1)\odot f(\mathcal{R}_{loc}, \varTheta_2),
\label{equ:7}
\end{equation}
where $\varTheta_1,\varTheta_2$ are learnable weights in FC-layers and $\odot$ denotes dot product operation.
\subsubsection{Enhanced Recognition using Location-based GGNN}
To model the potential relevance of $predicates$, we construct a location-based GGNN to improve the relationship representation $\mathcal{R}_{prm}$. Firstly, we give a brief introduction to the structure of GGNN.

In GGNN, the graph is constructed as $\mathcal{G} = (\bm V,\bm A)$, where $\bm V$ denotes a node set and $\bm A$ denotes an edge set. The detail propagation process is defined as
\begin{equation}
\begin{aligned}
&\bm h_v^{(1)}=[\bm x_v^\top,\bm 0]^\top,\\
&\bm a_v^{(t)} = \bm A_v^\top[\bm h_1^{(t-1)\top}\cdots \bm h_{|V|}^{(t-1)\top}]^\top + \bm b,\\
&\bm h_v^{(t)} = GRU(\bm h_v^{(t-1)},\bm a_v^{(t)}),
\end{aligned}
\label{equ:8}
\end{equation}
where $GRU$ is the Gated Recurrent Unit\cite{cho2014learning}. Firstly, for each node $v\in \bm V$, we initialize the node features $h_v^{(1)}$ with our input data $\bm x$. Then, during each step $t$, the features of neighborhood nodes (defined by $\bm A$) and itself hidden features update the node state with $GRU$ function. 

In this work, each node $v$ denotes a kind of $predicate$, thus $|\bm V|$ is equal to the numbers of defined $predicate$ in visual relationship dataset. To initialize the nodes, a linear operation is applied to transform the relationship representation $\mathcal{R}_{prm}\in\mathbb{R}^{m}$ into $\mathcal{R}_{pred}\in\mathbb{R}^{|\bm V|}$, where each bit links to one node. 
Due to the adjacent matrix $\bm A$ predefined as prior, we should construct $\bm A$ to interconnect the similar predicate nodes as the neighborhood nodes, which models the relevance of $predicates$. In this work, we observe there are two potential connections among $predicates$. One is that the phrase type $predicates$ have potential relationships with their every component $predicate$. For example, "walk next to", "walk" and "next to" exist potential relevance although they belong to independent categories. Another is that many $predicates$ exist relative location tendency. For example, "above", "ride" and "cover" exist implied relative location of "on". However, this kinds of potential relevance is too ambiguous to the artificial definition. In our paper, a novel and visualizing method is proposed to quantify them.

\begin{figure}[htb]
	\centering
	\includegraphics[width=7cm]{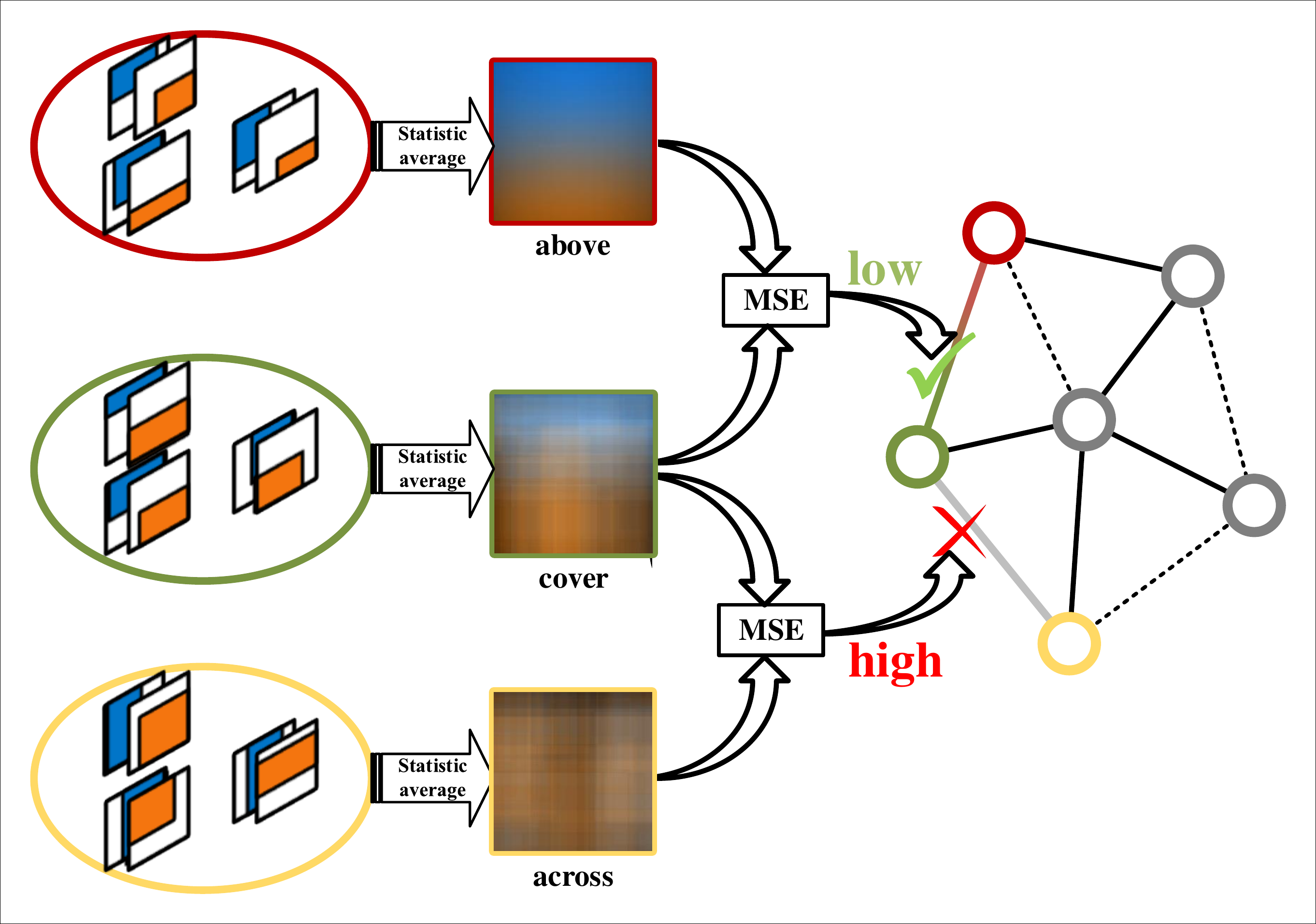}
	\caption{The illustration of the connection operation among different predicates. To visualize clearly, the \textcolor[rgb]{0,0.35,0.79}{blue} (resp. \textcolor[rgb]{0.96,0.46,0}{orange}) mask denotes the area of \textcolor[rgb]{0,0.35,0.79}{subject} (resp. \textcolor[rgb]{0.96,0.46,0}{object}). When the similarities (measured by MSE) of location anchors are below a thresh, nodes are interconnected in our location-based GGNN.}
	\label{fig:3}
\end{figure}

Based on the annotation bounding-boxes in the training set, we average the location masks that consist of two binary masks for each $predicate$. Shown in Figure~\ref{fig:3}, the blue mask represents the subject binary mask and the orange mask represents object binary mask. For each $predicate$, average operation is implemented among the same categories. Noteworthy, to only measure the relative location tendency, the binary masks only contain union regions of object-pairs and are resized to the same size in advance. Finally, there are $|\bm V|$ location anchors as prior, which preserve alignments with relative location tendency. In this way, if $predicates$ don't exist location tendency, there are no structural characteristics in the corresponding location anchors. like "across" in Figure~\ref{fig:3}. Based on these location anchors, similarities among $predicates$ are measured with mean-square error (MSE)
\begin{equation}
MSE(\mathcal X,\hat{\mathcal X})= \frac{1}{2}\sum_{n}\sum_{i,j}(\mathcal X_n(i,j)-\hat{\mathcal X}_n(i,j))^2, n \in\{sub,ob\},
\label{equ:9}
\end{equation}
where $\mathcal X,\hat{\mathcal X}$ are two location anchors and $(i, j)$ denotes the positions of each pixel. Nodes are interconnected when corresponding MSE is below a thresh as a hyper-parameter.

Based on above two hypotheses, every row of $\bm A$ is then normalized. Finally, the relationship representations are enhanced with the defined $\bm A \in\mathbb{R}^{|\bm V|\times|\bm V|}$. Following\cite{li2015gated}, the graph output of GGNN, $\mathcal{R}_{ggnn} \in \mathbb{R}^{|\bm V|} $ are defined as

\begin{equation}
\begin{aligned}
&o_v = O(\bm h_v^\top,\mathcal{R}_{pred,v}), v=1,2,\cdots, |\bm V|,\\
&\mathcal{R}_{ggnn} =[o_1,\cdots,o_{|\bm V|}],
\end{aligned}
\label{equ:10}
\end{equation}
where $O$ is an output network implemented by a FC-layer. 

The predicate recognition module produces the probabilities for $predicates$
\begin{equation}
P_{prm}= \mathrm {Softmax}(\mathcal{R}_{pred} \oplus \mathcal{R}_{ggnn}),
\label{equ:11}
\end{equation}
where $ \oplus $ is the element wise sum operation. 
With ea.~\ref{equ:10} and ~\ref{equ:11}, relative location based similarity information can be mined to model the relevance among $predicates$. Thus, those ambiguous but reasonable predictions can be clustered and smoothed to enhance predicate recognition. 
Finally, the training loss of predicate recognition module is

\begin{equation}
\mathcal{L}_{prm}= \frac{1}{N}\sum_{n=1}^N \mathcal{L}_{CEL}(P_{prm}^n,y_{prm}^n) ,
\label{equ:12}
\end{equation}
where $y_{prm}$ is the label for predicate recognition module, and $\mathcal{L}_{CEL}$ is Cross Entropy Loss.

\begin{table*}[!thp]
	\caption{Comparison of our proposed model with state-of-the-art methods on VRD testing set.}
	\label{tab:2}
	\setlength{\tabcolsep}{5.5pt}
	{
		\begin{tabular}{cccccccccccc}
			\toprule
			&\multicolumn{3}{c}{\bf Predicate Detection}&\multicolumn{4}{c}{\bf Phrase Detection}&\multicolumn{4}{c}{\bf Relationship Detection}\\
			Model 	&R@100/50,&R@100,&R@50,& R@100,&R@50,& R@100,& R@50,& R@100,&R@50,& R@100,& R@50, \\
			&k=1&k=70&k=70& k=1 & k=1 & k=70 & k=70& k=1 & k=1& k=70 & k=70 \\
			\midrule
			VTransE \cite{zhang2017visual}& 44.76 & - & -& 22.42 & 19.42 & -&-&15.20&14.07&-&-   \\
			Language-Pri \cite{lu2016visual}& 47.87 & 84.34 & 70.97& 17.03 & 16.17 & 24.90&20.04&14.70&13.86&21.51&17.35   \\
			TCIR \cite{zhuang2017towards}& 53.59 & - & - & 25.26 & 23.88 & -&-&23.39&20.14&-&-   \\
			CDD-Net \cite{cui2018context}& - & 93.76 & 87.57 &-&-&-&-&-&-&26.14&21.46 \\
			VIP \cite{li2017vip}&-&-&-&-&-  & 27.91 & 22.78&-&-& 20.01&17.31   \\
			DR-Net \cite{dai2017detecting}& - & 81.90 & 80.78  &-&-&23.45&19.93&-&-&20.88&17.73 \\
			VRL \cite{liang2017deep}&-&-&- & 22.60 & 21.37&-&-&20.79 & 18.19&-&-   \\
			Factorizable Net \cite{li2018factorizable}&-&-&-&-&-& 30.77 &26.03&-&-&21.20&18.32  \\
			LKD \cite{yu2017visual}& 55.16 & \underline{94.65} & 85.64& 24.03 & 23.14 & 29.43&26.32&21.34&19.17&\underline{31.89}&\underline{22.68}  \\
			STL \cite{chen2019soft}& - & - & -  &\underline{33.48}&\underline{28.92}&-&-&\underline{26.01}&\underline{22.90}&-&- \\
			Zoom-Net \cite{yin2018zoom}&\underline{55.98} & 94.56 & \underline{89.03}& 28.09 &24.82& \underline{37.34}&\underline{29.05}&21.41&18.92&27.30&21.37  \\
			\midrule
			RLM(ours)& \bf57.19 &\bf 96.48 &\bf 90.00 & \bf39.74 &\bf 33.20 &\bf 46.03&\bf 36.79&\bf 31.15&\bf 26.55&\bf 37.35&\bf30.22 \\
			\bottomrule
		\end{tabular}
	}
\end{table*}
\section{experiments}
Our experiments are based on pytorch\footnote{The scource code will be released on https://github.com/zhouhaocv/RLM-Net}, and the detection module follows the official released project\cite{massa2018mrcnn}.

\subsection{Training and Inference procedures}
In the training stage, our framework is trained in two stages. The first stage is to optimize the object detection module and our proposed location-guided object-pairs rating module jointly. Thus, the first stage training loss is defined as:

\begin{equation}
\mathcal{L}_1= \mathcal{L}_{odm} + \lambda \mathcal{L}_{orm} ,
\label{equ:13}
\end{equation}
where $ \mathcal{L}_{odm}$ is the loss for object detection module, which just follows the loss in \cite{massa2018mrcnn}. 

In the inference stage, three modules collaborate as a whole network to produce the final visual relationship predictions directly. Different from previous frameworks, in order to sort the reserved object-pairs proposals, object-pairs proposals scores $\tilde{\bm s}(sub,ob)$ are assigned into the final predictions. Therefore, given an object-pair $\langle b_i,b_j\rangle$ in one image, the visual relationship detection probabilities of all $predicates$ for $\langle sub,ob\rangle$ are 
\begin{equation}
\mathcal{\bm P}(sub,ob|{\langle b_i,b_{j}\rangle})= \hat{P}(sub|b_i)\cdot \hat{P}(ob|b_j)\cdot P_{prm}\cdot \tilde{\bm s}(sub,ob).
\label{equ:14}
\end{equation}

There are some implementation details in our experiments. Considering the good performance of Resnet-50-FPN\cite{lin2017feature}, we adopt the same structure to construct our backbone network. In the first training stage, $\lambda$ is set as 1. Then, in the second stage, the predicate recognition module is optimized with the frozen network shared with previous stage. In i-NMS, object-pairs numbers($N_o$) are 110, and threshold for NMS($N_t$) is 0.25. The relationship representation($R_{prm}$) is 128 dims. In GGNN, nodes are interconnected when corresponding MSE is below 1000 for VRD (500 for VG).

\subsection{Evaluation Metrics}
Because annotations of visual relationship are not exhaustive, mAP evaluation metrics will penalize positive predictions which are absent in ground truth. We follow \cite{lu2016visual} to use Recall@50 (R@50) and Recall@100 (R@100) as our evaluation metrics. R@n computes the Recall using the top n object-pairs proposals' predictions in one image. Following \cite{yu2017visual}, we also set a hyperparameter k, which means to take the top k predictions into consideration per object-pair. 
In visual relationship detection task, R@n,k=1 is equivalent to R@n in \cite{lu2016visual}. R@n,k=70 in VRD and R@n,k=100 in VG is equivalent to take all $predicates$ into consideration. In visual relationship detection task, there are three evaluation metrics, including predicate detection, phrase detection and relationship detection. Because the object-pairs proposals are given in the inference stage, only the performances of predicate recognition module are evaluated in predicate detection evaluation. Comparing to predicate detection metric, phrase detection (that needs to detect union regions of object-pairs) and relationship detection (that needs to detect objects respectively) evaluate the performances of entire visual relationship detection framework.

\subsection{Experiments on Visual Relationship Detection}
We evaluate our model in Visual Relationship Detection dataset \cite{lu2016visual}, which contains 70 predicates and 100 objects. The whole dataset is split into 4000 images for training and 1000 images for testing.

\subsubsection{Comparison to state-of-the-art Methods}

We compare our proposed model that denotes "$RLM(ours)$" with some related methods \cite{zhang2017visual,lu2016visual,zhuang2017towards,chen2019soft,li2017vip,dai2017detecting,liang2017deep,li2018factorizable,yu2017visual,yin2018zoom,cui2018context} on three metrics in Table~\ref{tab:2}. In the object-pairs proposing stage, we only reserved 110 object-pairs proposals in this experiment. As the results shown in Table~\ref{tab:2}, comparing to existing methods, our proposed model achieves significant improvements in all evaluation metrics. Especially in the most challenging metric, relationship detection evaluation, our model achieves more gains, e.g. 31.15\% vs 26.01\% for R@100,k=1. We contribute our improvements to two main fold. One is our proposed improved predicate recognition module. Through multiple modal features fusion and introduction of statistic relative location prior in GGNN, our predicate recognition module achieves state-of-the-art performances on the corresponding evaluation metric "predicate detection". Another one is our proposed novel object-pairs proposing scheme, which not only select valid object-pairs effectively but also sorts them in final relationship predictions. We also compare our model with some existing methods on "Zero-shot Set" in Table~\ref{tab:3}. "Zero-shot set" removes relationship triplets that already exist in the training set. Thus predicate detection metric in Zero-shot set can be used to measure the generalization ability of specific predicate recognition module. The improvements mean our model also has a better generalization performance in some unseen triplets.
\begin{table}[!htb]
	\caption{Comparison of our proposed model with the state-of-the-art methods on Predicate detection evaluation in VRD Zero-shot Set.}
	\label{tab:3}
	\setlength{\tabcolsep}{7pt}
	{
		\begin{tabular}{cccccc}
			\toprule
			\multirow {2}{*}{Model}
			&R@100,&R@50,&R@100,&R@50, \\
			&k=1&k=1 & k=70 & k=70 \\
			\midrule
			Language-Pri \cite{lu2016visual}& 8.45& 8.45 & 50.04 & 29.77   \\
			TCIR \cite{zhuang2017towards}& 16.42& 16.42 & - & -   \\
			Weakly-sup \cite{peyre2017weakly}&\underline{23.6} &\underline{23.6} & - & -   \\
			LKD \cite{yu2017visual}& 16.98 & 16.98 & \underline{74.65} &\underline{54.20}   \\
			\midrule
			RLM(ours)& \bf24.64& \bf24.64 &\bf 88.11 &\bf 72.03\\
			\bottomrule
		\end{tabular}
	}
\end{table}
\subsubsection{Component Analysis} 
\begin{figure*}[!thb]
	\centering
	\includegraphics[width=17.8cm]{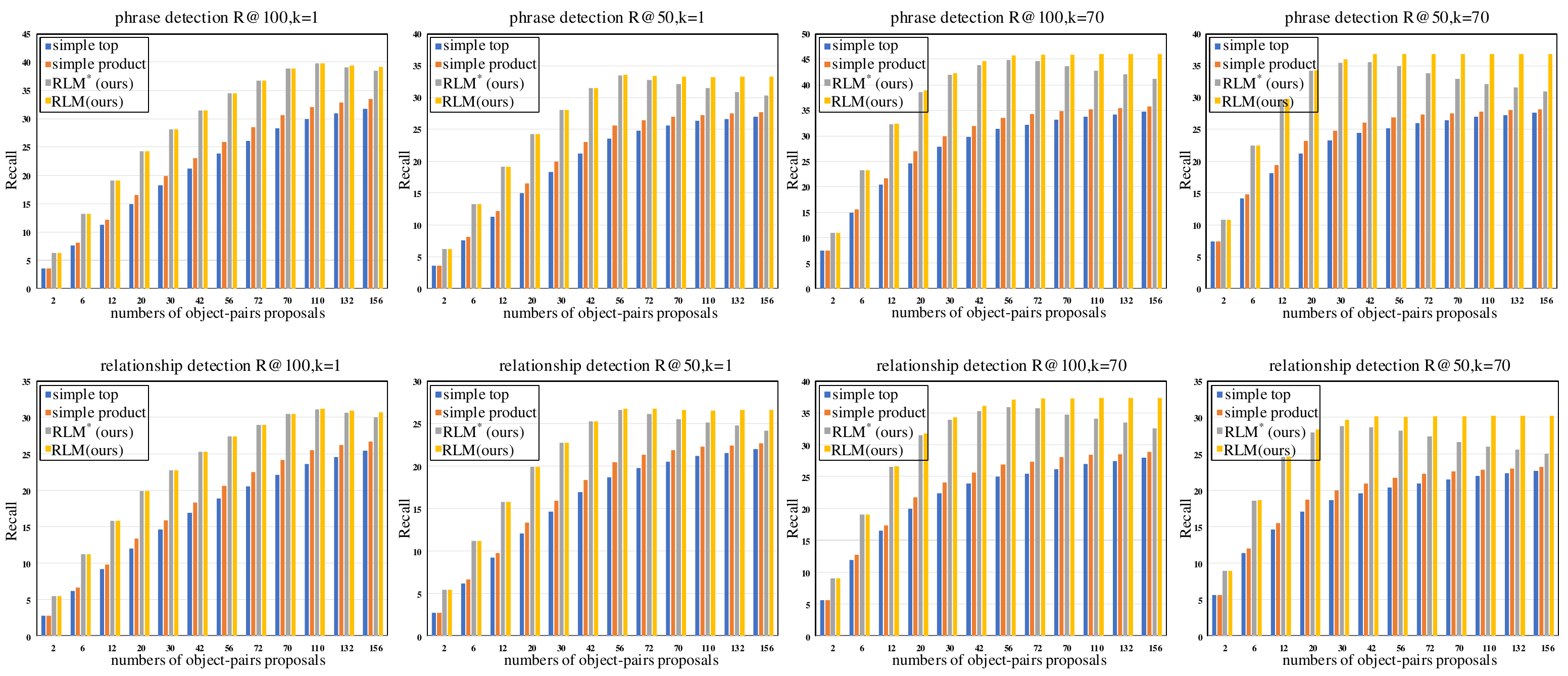}
	\caption{Comparison analysis of our object-pairs proposing scheme with other popular proposing methods. The top row is the results on the phrase detection metric, the second row is the results on the relationship detection metric. The X-axis denotes the number of reserved object-pairs proposals, and Y-axis denotes Recall values.}
	\label{fig:4}
\end{figure*}
In our work, we follow the existing method to construct our object detection module directly, and the analysis of our proposed components keep the same settings in the object detection module.

Firstly, to explore the influences of different components in our predicate recognition module, we evaluate different setting combinations on predicate detection. All of the results are listed in Table~\ref{tab:1}. Comparing to "Zero-shot set", "Entire set" denotes the entire testing set. Three kinds of settings are analyzed in the predicate recognition module. 1) The first one is combinations of different modal features. The input of our model contains language priors, relative location information and visual features. "$vis$" denotes that only visual features are used, "$vis+language$" denotes that language prior and visual features are fused, and "$vis+location$" denotes relative location information and visual features are fused. Both "$vis+location$" and "$vis+language$" outperform "$vis$", which proves that only visual features are limited to capture the subtle interactions of object-pairs. 
2) The second one is the different combination ways of three modal features. "$concate$" means the concatenation operation, "$average$" means the average operation. In our experiment, we find product operation ("$muti+product$") achieves best performance. We consider product operation makes gradient backward into the three branches easily comparing to another two operations so as to achieve better multimodal features fusion. 
3) The third one is the influences of our location-based GGNN in predicate recognition. Comparing to "$multi+product+GGNN$", "$multi+product$" is not including location-based GGNN. In the experiment, "$multi+product+GGNN$" outperforms "$multi+product$" in most prediction detection metrics, especially in R@n,k=70. It proves that our proposed location-based GGNN is effective to model the potential spatial relevance among different $predicates$ so as to improve predicate recognition performances.

\begin{table}[!htb]
	\caption{
		Performances of our predicate recognition module in different settings on the predicate detection evaluation.
	}
	\label{tab:1}
	\setlength{\tabcolsep}{1.5pt}
	{
		\begin{tabular}{ccccc}
			\toprule
			&\multirow {2}{*}{Model}
			&R@100/50,&R@100,&R@50, \\
			& &k=1 & k=70 & k=70 \\
			\midrule
			\multirow {7}{*}{Entire Set}
			&vis  & 33.74& 86.06 &72.74 \\
			&vis+language & 53.78 & 94.68 & 86.96  \\
			&vis+location & 48.83 & 92.71 & 83.71   \\
			\cmidrule{2-5}
			&multi+concate& 56.39 & 94.44 & 88.35   \\
			&multi+average& 52.28 & 94.93 & 88.79  \\
			&multi+product& 56.81 & 95.97 & 89.33  \\
			\cmidrule{2-5}
			&multi+product+GGNN&\bf 57.19 &\bf 96.48 & \bf90.00  \\
			\hline\midrule
			\multirow {7}{*}{Zero-shot Set}
			&vis  & 12.83& 73.57 &53.29  \\
			&vis+language & 19.33 & 80.92 & 61.76 \\
			&vis+location & 24.55 & 84.75 & 68.86   \\
			\cmidrule{2-5}
			&multi+concate& 23.70 & 85.29 & 71.34   \\
			&multi+average& 23.44 & 84.94 & 71.77  \\
			&multi+product& \bf24.72 & 87.17 & 71.34  \\
			\cmidrule{2-5}
			&multi+product+GGNN& 24.64 &\bf 88.11 &\bf 72.03  \\
			\bottomrule
		\end{tabular}
	}
\end{table}

Secondly, to explore the performance of our object-pairs proposing scheme, we compare the influences of the reserved numbers of object-pairs proposals in ours and other two popular methods. For the sake of fairness, all of the proposing schemes are based on the same predicate recognition module "$multi+product+GGNN$". The results are shown in Figure~\ref{fig:4}. "$simple\,top$" denotes the earliest object-pairs proposed method, which reserves objects based on the objectiveness scores directly. "$simple\,product$" denotes that specific object-pairs are reserved based on the product of $\langle sub-ob\rangle$ objectiveness scores. "$RLM^*(ours)$" adopts our proposing scheme in object-pairs proposing stage, but object-pairs proposals scores $\tilde{\bm s}(sub,ob)$ are not assigned in final visual relationship predictions. "$RLM(ours)$" is our complete proposing scheme. The top row is the results on the phrase detection metric; the second row is the results on the relationship detection metric. From Figure~\ref{fig:4}, we can see our proposing scheme achieves significant improvements compared with traditional methods. With numbers of reserved proposals increasing, the performances of both "$RLM^*(ours)$" and "$RLM(ours)$" improve rapidly. That means our proposing scheme can reserve meaningful object-pairs as much as possible. Generally, when only obtained a few object-pairs proposals (about less than 100 proposals), "$RLM^*(ours)$" and "$RLM(ours)$" have similar performances. Interestingly, with the reserved numbers increasing continuously, the recall of "$RLM^*(ours)$" has a slight drop comparing to "$RLM(ours)$". This is because too many proposals reserved will cause inevitable redundancy of object-pairs proposals. To alleviate this issue, we introduce the object-pairs proposals scores $\tilde{\bm s}(sub,ob)$ into our final predictions denoted as "$RLM(ours)$" to sort the reserved object-pairs. In this way, more meaningful object-pairs can be assigned higher scores in the final predictions, which makes our model insensitive to the reserved number of object-pairs proposals.

\subsection{Experiments on Visual Genome}
Visual Genome(VG)\cite{krishna2017visual} is a large scale relation dataset that exists much annotation noise. Thus, we adopt a clean subset\cite{zhang2017visual} where an official pruning are applied. As a clean subset, there are 99658 images with 200 object categories and 100 predicates. The whole dataset is split into 73,801 images for training and 25,857 images for testing.

\begin{table}[!htp]
	\caption{Comparison of our proposed model with the state-of-the-art methods on VG testing set. }
	\label{tab:4}
	\setlength{\tabcolsep}{2pt}
	{
		\begin{tabular}{cccccccc}
			\toprule
			&\multicolumn{2}{c}{\bf Predicate}&\multicolumn{2}{c}{\bf Phrase}&\multicolumn{2}{c}{\bf Relationship}\\
			Model&\multicolumn{2}{c}{\bf Detection}&\multicolumn{2}{c}{\bf Detection}&\multicolumn{2}{c}{\bf Detection}\\
			&R@100&R@50& R@100&R@50& R@100& R@50 \\
			\midrule
			Visual phrase \cite{sadeghi2011recognition}& - & - &4.27& 3.41 & - & -  \\
			VTransE \cite{zhang2017visual}& 62.87 & 62.63 & 10.45& 9.46 & 6.04 &5.52  \\
			Shuffle \cite{yang2018shuffle}& 62.94 &62.71 & - & - & - & -   \\
			VSA-Net \cite{han2018visual}& 64.53 &\underline{64.41} & 9.97 & 9.72 & 6.28 & 6.02   \\
			PPR-FCN \cite{zhang2017ppr}&\underline{64.86}&64.17&11.08&10.62&6.91  & 6.02   \\
			DSL \cite{zhu2018deep}& - & - & 15.61 &13.07&8.00&6.82 \\
			STL \cite{chen2019soft}& - & - & \underline{18.13}  &\underline{14.62}&\underline{9.41}&\underline{7.93} \\
			\midrule
			Simple top& - & - & 19.21 &17.19 &12.04 &10.87 \\
			Simple product& - & - &20.48 &17.94 &12.96 &11.42 \\
			$\mathrm {RLM}^\dagger$(ours)&68.20 &67.93 &33.28 &26.08 &20.78 &16.65 \\
			RLM(ours)& \bf69.87 &\bf 69.57 &\bf 33.92 & \bf26.60 &\bf21.17 &\bf 16.96 \\
			\midrule\midrule
			k=100,RLM(ours)& 95.40 &89.80 &35.00 & 27.52 &22.65 &18.19 \\
			\bottomrule
		\end{tabular}
	}
\end{table}

We compare our complete model denoting "$RLM(ours)$" with some existing methods. Most existing methods only evaluate on R@n,k=1 metric in VG, and thus the results listed in Table~\ref{tab:4} are all evaluated on the same metric. From the results, we can see that our proposed model still achieves state-of-the-art performances on all of the metrics. Especially, the performances of our model outperform more than twice the existing methods on phrase detection and relationship detection. To further explore the influences of different components, we make some extra experiments. Based on our predicate recognition module, "$simple\,top$" and "$simple\,product$" denote two existing object-pairs proposing schemes mentioned in the previous section. "${RLM}^\dagger(ours)$" denotes that the location-based GGNN are removed compared to "$RLM(ours)$". As shown in Table~\ref{tab:4}, the huge gains prove the effectiveness of our proposed visual relationship detection model. 
To the need of further research, we also experiment our model on R@n,k=100 listed as "$k=100,RLM(ours)$", which means all of the predictions are evaluated.
\section{conclusion}
In our work, we exploit the relative location information to guide visual relationship detection in two stages and construct an improved relationship detection framework. 
We propose a location-guided object-pairs rating module to model the probability of object-pairs interconnected. Specific object-pairs can be reserved and sorted effectively in our object-pairs proposing stage. In order to improve the performances of predicate recognition further, we propose a location-based GGNN to model the relevance among different $predicates$. 
The experiment on VRD shows that our proposed framework achieves state-of-art performances, and even outperforms more than twice existing methods for phrase detection and relationship detection in a large scale dataset VG.  

\section{Acknowledgments}
This work was partly funded by NSFC (No.61571297), the National Key Research and Development Program (2017YFB1002400), and STCSM (18DZ1112300).

\bibliographystyle{ACM-Reference-Format}
\balance
\bibliography{sample-base}

\end{document}